# The Collaboration Paradox: Why Generative AI Requires Both Strategic Intelligence and Operational Stability in Supply Chain Management

**Soumyadeep Dhar**
Department of Industrial & Systems Engineering
Indian Institute of Technology, Kharagpur (IIT-KGP)
Kharagpur, West Bengal, India
sdhar1602@kgpian.iitkgp.ac.in

## Abstract

The rise of autonomous, AI-driven agents in economic settings raises critical questions about their emergent strategic behavior. This paper investigates these dynamics in the cooperative context of a multi-echelon supply chain, a system famously prone to instabilities like the bullwhip effect. We conduct computational experiments with generative AI agents, powered by Large Language Models (LLMs), within a controlled supply chain simulation designed to isolate their behavioral tendencies. Our central finding is the "collaboration paradox": a novel, catastrophic failure mode where theoretically superior collaborative AI agents, designed with Vendor-Managed Inventory (VMI) principles, perform even worse than non-AI baselines. We demonstrate that this paradox arises from an operational flaw where agents hoard inventory, starving the system. We then show that resilience is only achieved through a synthesis of two distinct layers: high-level, AI-driven proactive policy-setting to establish robust operational targets, and a low-level, collaborative execution protocol with proactive downstream replenishment to maintain stability. Our final framework, which implements this synthesis, can autonomously generate, evaluate, and quantify a portfolio of viable strategic choices. The work provides a crucial insight into the emergent behaviors of collaborative AI agents and offers a blueprint for designing stable, effective AI-driven systems for business analytics.

## 1. Introduction
The rapid integration of Generative Artificial Intelligence (GenAI), particularly Large Language Models (LLMs), into business decision-making is opening a new frontier for algorithmic automation in complex domains such as supply chain management (Wang et al., 2024). Unlike traditional optimization systems, LLM-based agents can reason over large volumes of heterogeneous and unstructured information, potentially enabling more adaptive and intelligent operational decision-making (Agrawal, 2024; Touvron et al., 2023). These capabilities have generated growing interest in deploying AI agents to assist or automate strategic and operational tasks across supply chain networks.

At the same time, the increasing autonomy of AI agents raises fundamental questions regarding their emergent strategic behavior. Recent research has shown that algorithmic agents, from reinforcement learning systems to modern AI models, can autonomously learn supracompetitive or collusive strategies in pricing environments without explicit instructions to do so (Calvano et al., 2020; Fish et al., 2026). While such findings are typically discussed in competitive market contexts, they also raise an important but less explored question: *how will autonomous AI agents behave in cooperative operational systems?*

Supply chains represent a particularly important setting for examining this issue. Although supply chains are designed to operate cooperatively to minimize overall system costs, they are typically characterized by decentralized decision-making, imperfect information sharing, and misaligned incentives. These conditions often produce systemic instability, the most well-known manifestation of which is the bullwhip effect, where small demand fluctuations at the retail level amplify into large inventory oscillations upstream (Lee et al., 1997; Sterman, 1989). The central research question, therefore, is not simply "*Can an LLM generate a supply chain strategy?*" but rather, "*Can a system of multiple, interacting AI agents execute strategies in a way that enhances stability, or does their automated interaction inadvertently create new and even more volatile failure modes?*" The literature on the brittleness and unforeseen consequences of AI systems suggests this is a non-trivial risk (Raji et al., 2021).

To investigate this problem, this study conducts a series of computational experiments within a controlled three-echelon supply chain simulation environment designed to reproduce bullwhip dynamics, similar in spirit to the Beer Game widely used in operations management research and education (Oroojlooyjadid et al., 2020). Within this environment, we evaluate progressively more sophisticated decision-making agents, beginning with a non-AI baseline, followed by a Retrieval-Augmented Generation (RAG)-enabled decision agent, and finally a collaborative multi-agent framework inspired by Vendor-Managed Inventory (VMI) principles.

Our findings reveal a striking and counterintuitive result that we term the "collaboration paradox." Contrary to expectations, the most advanced collaborative AI framework consistently underperformed relative to simpler models, producing service levels even worse than the non-AI baseline. Further investigation shows that this failure is not due to implementation error but emerges from the interaction dynamics of the agents themselves. Specifically, strategically sound policies generated by the AI led to flawed operational execution that produced a hoarding effect, a previously underexplored instability analogous to the bullwhip effect.

Through iterative experimentation and model refinement, we demonstrate that system stability can only be achieved by integrating two complementary layers of intelligence: (1) high-level strategic policy generation, where AI agents establish robust operational targets, and (2) low-level collaborative operational protocols that coordinate real-time execution across agents. The resulting architecture illustrates how strategic reasoning and operational coordination must be jointly designed to ensure stable outcomes in AI-driven supply chain systems.

This study contributes to the growing literature on AI-enabled operations management in two key ways. First, it provides a replicable experimental framework for studying emergent behaviors in multi-agent AI systems operating within supply chain environments. Second, it offers a data-driven cautionary perspective on the deployment of GenAI in complex business systems, demonstrating that intelligent decision agents alone are insufficient without robust system-level design.

The remainder of the paper is organized as follows. Section 2 reviews the literature on supply chain resilience and AI applications in operations management. Section 3 describes the simulation environment and the architecture of the proposed agent-based models. Section 4 presents the experimental results across the different agent configurations. Section 5 discusses the implications of these findings for AI-driven supply chain management. Finally, Section 6 concludes the study and outlines directions for future research.

## 1.1 Objectives

The primary objective of this study is to examine how autonomous Generative AI agents behave when deployed within decentralized supply chain environments and to determine whether their interactions enhance or destabilize system performance. In particular, the research focuses on understanding how AI-driven decision-making influences operational dynamics in multi-echelon supply chains that are naturally prone to instability due to decentralized information and local optimization incentives.

To address this objective, the study develops a controlled computational simulation designed to replicate the dynamics of a multi-echelon supply chain and reproduce conditions under which the bullwhip effect typically emerges. Within this experimental environment, a series of progressively sophisticated decision-making frameworks are implemented and evaluated, beginning with a non-AI baseline and extending to Retrieval-Augmented Generation (RAG) enabled agents and collaborative multi-agent architectures inspired by Vendor-Managed Inventory (VMI) principles. By systematically comparing these approaches, the research aims to evaluate how different forms of AI-driven decision logic affect operational performance, including service levels and inventory stability across the supply chain.

Beyond performance evaluation, the study also seeks to identify the emergent behavioral patterns that arise when multiple autonomous agents interact within decentralized operational structures. In doing so, it aims to uncover potential failure modes, such as amplified bullwhip dynamics or inventory hoarding, that may arise from seemingly rational AI-generated strategies. Ultimately, the goal is to derive practical design insights for building stable AI-enabled supply chain systems, highlighting the importance of integrating high-level strategic reasoning with robust collaborative operational protocols.

## 2. Literature Review

Our research is situated at the confluence of three distinct yet increasingly intertwined academic domains: supply chain resilience, the application of artificial intelligence in operations management, and the nascent field of generative AI-driven business analytics. This review synthesizes key contributions from each area to contextualize our work and highlight its novelty.

### 2.1 Supply Chain Resilience and Disruption Management

The study of supply chain resilience has evolved significantly from its early focus on redundancy and efficiency. Foundational work identified key sources of risk, including operational failures, natural disasters, and geopolitical instability, and proposed static mitigation strategies such as holding safety stock, diversifying the supplier base, and building flexible production capacity (Chopra and Sodhi, 2004; Tang, 2006). While effective against known risks, these strategies are often economically burdensome and have been shown to be insufficient for coping with unexpected, high-impact “black swan” events (Taleb, 2007; Sheffi, 2005). This has led to a paradigm shift towards viewing resilience not merely as robustness (the ability to withstand a shock) but as agility and adaptability (the ability to recover and dynamically respond) (Christopher and Peck, 2004; Ponomarov and Holcomb, 2009).

More recent research has focused on developing quantitative models for resilience, often employing techniques from network theory and system dynamics to understand how disruptions propagate through complex supply networks (Klibi et al., 2010). Scholars have developed metrics to quantify resilience, such as the time-to-recover (TTR) and performance impact, providing a more rigorous basis for evaluating different network structures and recovery policies (Hosseini et al., 2019). However, the majority of these models assume that response strategies are either pre-defined or can be optimized using traditional operations research (OR) methods. They do not typically account for the challenge of generating novel strategies in real-time based on incomplete and unstructured information, a critical gap that our work aims to address.

### 2.2 Artificial Intelligence in Operations Management

The application of Artificial Intelligence (AI) and Machine Learning (ML) to operations management (OM) is a well-established field. Early contributions focused heavily on predictive analytics, using techniques like time-series analysis, regression, and neural networks to improve demand forecasting, predict machine failures, and assess supplier risk (Carbonneau et al., 2008; Makridakis et al, 2018). These predictive models have provided significant value by improving the quality of the data fed into traditional planning and optimization systems.

More recently, the focus has expanded to include prescriptive analytics, where AI systems recommend specific actions. Reinforcement Learning (RL), for instance, has been successfully applied to dynamic inventory control and pricing problems, allowing agents to learn optimal policies through trial-and-error in a simulated environment (Wu and Sekiguchi, 2019; Silver et al., 2017). While powerful, RL approaches typically require a well-defined state-action space and a clear reward function, which can be difficult to formulate for complex, strategic problems like disruption response. Furthermore, both predictive and RL-based systems often operate as “black boxes,” lacking the explainability and reasoning capabilities necessary for high-stakes strategic decision-making in a business context (Adadi and Berrada, 2018; Dhar and Goh, 2025).

### 2.3 Generative AI and LLMs as a New Frontier in Business Analytics

The advent of Generative AI and LLMs marks a potential inflection point for AI in OM. Unlike previous AI paradigms, LLMs excel at understanding and reasoning over unstructured text, allowing them to ingest and synthesize human knowledge from documents, reports, and real-time news feeds (Agrawal et al., 2023). This capability has opened up a new research frontier focused on creating AI agents that can act as strategic partners and decision-support systems.

Initial exploratory studies have demonstrated the potential of LLMs to perform a variety of business-related tasks, including drafting marketing copy, summarizing financial reports, and generating business process models Davenport and (Mittal, 2023; II and Katz, 2022). In the supply chain domain, researchers have begun to explore using LLMs as "co-pilots" that can answer complex queries or act as components within larger agent-based systems (Wang et al., 2024; Li et al., 2023). For example, the RAG technique allows an LLM to ground its reasoning in a specific, curated knowledge base, mitigating the risk of hallucination and enabling context-specific problem-solving (Lewis et al., 2020).

Our work builds directly on this nascent frontier. Although existing studies have shown that LLMs can generate plausible strategies, the question of how a system of multiple LLM-based agents interacts and whether that interaction is stable or chaotic remains largely unexplored. Our research contributes to this conversation by moving beyond the capabilities of a single agent to empirically investigate the emergent dynamics of a multi-agent, collaborative AI system. By chronicling the journey from a brittle, unstable system to a resilient one, we provide critical, practical insights into the deep challenges and necessary conditions for successfully deploying generative AI in complex, real-world business analytics applications.

## 3. Methodology

Our research is conducted through a multi-stage, iterative process of computational modeling and experimentation. We first construct a simulated supply chain environment to serve as a digital testbed. Within this environment, we design and evaluate a series of increasingly sophisticated agent-based models, with each iteration revealing critical insights that inform the next. This evolutionary approach allows us to not only assess the performance of our final proposed framework but also to systematically identify the necessary conditions for a Generative AI-driven system to operate successfully in a complex, dynamic environment.

### 3.1 A Supply Chain Laboratory for Studying Agent Behavior

To isolate the strategic and behavioral dynamics of our AI agents, we designed a minimalist, three-echelon, single-product supply chain simulation. This environment is not intended to be a high-fidelity digital twin of a specific real-world operation, but rather a controlled "laboratory," analogous to the classic Beer Game simulation used extensively in operations management research and education (Oroojlooyjadid et al., 2020). The environment's parameters (e.g., lead times, demand variability) are specifically chosen to be sensitive to the bullwhip effect (Lee et al., 1997). This deliberate simplification allows us to transparently observe, analyze, and measure the emergent behaviors, such as instability, hoarding, and collaboration, that result from the agents' different policies, providing a clear and replicable testbed for our core research questions. The simulation is developed in pure Python to ensure full control over the logic and is run on a daily time step for a 150-day horizon.

### 3.2 Environment and Core KPIs

The supply chain consists of three entities: a Supplier with finite production capacity, a central Manufacturer, and a Retailer facing stochastic customer demand modeled as a Poisson process, $D \sim \text{Pois}(\lambda = 10)$. System performance is measured on two primary KPIs:

1. Total Supply Chain Cost ($): The sum of inventory holding costs for all entities and backorder penalty costs for unfulfilled customer demand.
2. 2. Customer Service Level (%): The overall fill rate, calculated as (Total Units Fulfilled/Total Demand).

### 3.3 Model Architectures and Iterative Development

Our methodology is defined by the benchmarking of three distinct models, which also represent the chronological journey of our research. Table 1 summarizes the key architectural differences between the models, which are detailed in the subsequent sections.

**3.3.1 Model 1: The Static Baseline:** This model represents a traditional, non-AI approach. All entities operate using a static Order-Up-To (OUT) inventory policy with a fixed, pre-determined target level. It serves as a clear performance benchmark.

**3.3.2 Model 2: The Selfish RAG Agent:** This model introduces a "selfish" intelligent agent at the Manufacturer node, endowed with a RAG-based SGA. The agent acts on local information to pursue its own goals, such as placing a large emergency order after a stockout, without visibility into the wider system.

Table 1: Specification of Benchmarked Agent-Based Models

| Component | Model 1: Baseline | Model 2: Selfish AI | Model 3: Collaborative AI |
|---|---|---|---|
| Inventory Policy | Static OUT | Static OUT | Proactive, AI-set OUT |
| Policy Setter | Human (Fixed) | Human (Fixed) | SGA at t=0 |
| Ordering Logic | Decentralized | Decentralized | Centralized (VMI) |
| Replenishment | N/A | N/A | Proactive Push |
| AI Capability | None | Reactive RAG (Selfish) | Proactive & Reactive RAG |

**3.3.3 Model 3:** The Final Collaborative Framework**:** Our final model represents the synthesis of lessons learned. It features a two-layer architecture where an SGA first acts as a proactive "Policy Advisor" to set intelligent, system-wide inventory targets. Execution is then handled by a collaborative VMI-style protocol where the Manufacturer manages a centralized ordering process and proactively pushes inventory downstream to the Retailer. This final architecture creates a system where strategic intelligence and stable operational execution work in harmony.

## 3.4 The "Strategic Choice" Experiment

Using our stable and successful final model (Model 4) as the foundation, we designed our main experiment to fully evaluate the strategic capabilities of the SGA, directly addressing our remaining research questions.

**3.4.1 SGA Strategy Generation (RQ2):** For a given disruption scenario (a transportation blockage), we enhanced the SGA's knowledge base to contain three distinct, viable strategies with varying trade-offs between cost and speed. To test the agent's ability to select the most relevant plan, we tasked the SGA with querying this knowledge base. The agent uses the Retrieval-Augmented Generation (RAG) framework for the retrieval step, performing a similarity search to find the most relevant strategic document. To ensure deterministic and reproducible results, the generation step was simulated: a dedicated Python function then parses the retrieved text to extract the pre-defined parameters of the strategy into a machine-readable format.

**3.4.2 Virtual Expert Panel Evaluation (H1):** To qualitatively assess the strategies retrieved by the SGA, we implemented a 'VirtualExpertPanel'. To maintain experimental control and avoid the variability of live LLM calls, this panel's evaluative reasoning was simulated. The simulated agent receives the portfolio of strategies and assigns a qualitative rating for each one (e.g., Cost Rating, Speed Rating) based on a set of programmatic rules that interpret the parameters and descriptions extracted from the knowledge base. For example, a strategy with a high 'transport cost premium' would be rated as "Very High" cost. This allows for an automated, repeatable, and transparent evaluation of the RAG agent's retrieval capabilities.

**3.4.3 Quantitative What-If Analysis (RQ3)**: Finally, to quantify the performance of each strategic option, we ran our full simulation three times, once for each of the strategies proposed by the SGA. By applying the parameters of each strategy within the simulation, we generated the final cost and service level KPIs for each choice, allowing for a clear analysis of the strategic trade-offs.

## 3.5 Disruption Scenarios

To test the robustness of our first three models (Baseline, Selfish, and the initial Collaborative attempt), we subjected them to a suite of four distinct disruption scenarios, each beginning at day 60 of a 150-day simulation:

- Supplier Failure: The Supplier's production time doubles for 20 days.
- Transportation Disruption: An additional lead time is added to all of the Manufacturer's inbound shipments for 15 days.
- Demand Surge: Customer demand at the Retailer increases by 50% for 20 days.
- Quality Failure: 70% of the Manufacturer's on-hand inventory is instantly wiped out at a single point in time

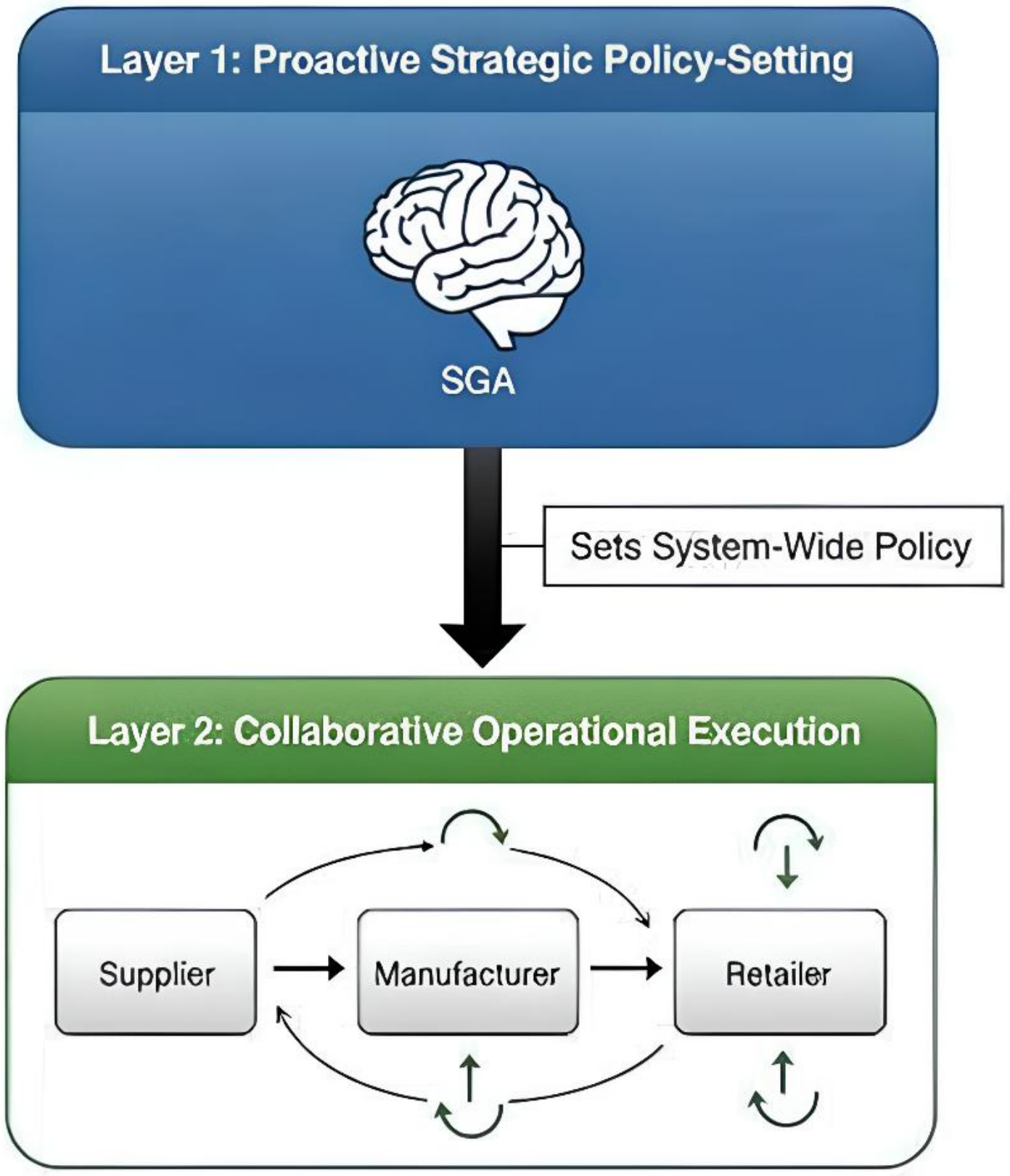

Figure 1: The Final Two-Layer Collaborative Framework (Model 4). Success was achieved by synthesizing a high-level strategic layer, where the SGA sets proactive system-wide policy, with a low-level operational layer that uses collaborative execution (VMI with downstream replenishment) to maintain stability.

## 4. Results

This section presents the quantitative and qualitative results of our computational experiments. We structure our findings to follow the iterative model development lifecycle detailed in the Methodology, beginning with the baseline models, proceeding through our insightful failures, and culminating in the successful performance of our final collaborative framework.

### 4.1 The Inevitable Failure of Non-Collaborative Models

Our initial set of experiments was designed to quantify the performance of traditional, noncollaborative supply chain policies when subjected to major disruptions. We benchmarked two models: a Static Baseline using a simple Order-Up-To (OUT) policy, and a Selfish RAG Agent where the Manufacturer could make locally-optimized, AI-driven decisions. Each model was run 30 times against four distinct disruption scenarios. The aggregate results are presented in Table 2.

The results are stark and unambiguous. Across nearly all scenarios, both the Static Baseline and the Selfish RAG models failed catastrophically, achieving an average customer service level of only ≈11%. This indicates that over 88% of customer demand went unfulfilled, confirming that simplistic operational policies are fundamentally incapable of creating a resilient supply chain. Notably, the introduction of a locally-optimized, “selfish” AI agent provided no significant benefit and, in some cases, slightly worsened performance.

The one exception is the Quality Failure scenario. Here, the Selfish agent’s ability to execute a reactive, large emergency order allowed it to recover a portion of its wiped-out inventory, leading to a modest but statistically significant improvement in both cost and service level. This isolated success highlights that while reactive AI can be beneficial for simple, localized shocks, it is not a solution for the systemic instabilities that plague the supply chain. The overall conclusion is clear: without collaboration, both naive and locally intelligent systems are brittle.

Table 2: Aggregate Performance of Non-Collaborative Models Across All Scenarios (Mean ± Std. Dev.)

| **Disruption Scenario** | **Model 1: Static Baseline** | | **Model 2: Selfish RAG Agent** | |
|---|---|---|---|---|
| | **Total Cost ($)** | **Service Level (%)** | **Total Cost ($)** | **Service Level (%)** |
| Supplier Failure | 13,697 ± 405 | 11.73 ± 0.52 | 13,756 ± 248 | 11.67 ± 0.32 |
| Transport Disruption | 13,791 ± 402 | 11.63 ± 0.51 | 13,769 ± 371 | 11.67 ± 0.46 |
| Demand Surge | 14,204 ± 305 | 10.98 ± 0.34 | 14,157 ± 295 | 11.04 ± 0.33 |
| Quality Failure | 13,763 ± 323 | 11.65 ± 0.39 | **9,921 ± 285** | **16.97 ± 0.63** |

## 4.2 The Paradox of Flawed Collaboration: A Chronicle of Failure

Our next stage of research focused on building a collaborative model based on Vendor-Managed Inventory (VMI) principles. This journey, however, was defined by a series of insightful implementation failures that revealed the deep logical requirements for a multi-agent system to function. Our initial collaborative models, despite being theoretically superior, consistently produced catastrophic failures, with service levels often falling below 5%.

The primary failure mode, as illustrated in the conceptual diagram in Figure 2, was the "hoarding effect." In our first VMI implementation, the Manufacturer agent would correctly place a single, consolidated order on behalf of the entire downstream chain. However, due to a flaw in its operational logic, it would receive and store all incoming inventory in its own warehouse without any mechanism to proactively push stock to the Retailer. The Retailer, now passive and no longer placing orders, was effectively starved of inventory. This led to a paradoxical situation where the Manufacturer's inventory skyrocketed while the Retailer's inventory plummeted, causing a complete system collapse at the customer-facing end. Subsequent iterations revealed further subtle bugs related to inventory accounting ('inventory position' tracking), each of which resulted in system failure. These experiments, while unsuccessful from a performance standpoint, were critical to our research, as they empirically demonstrated that a successful collaborative AI system requires not just shared information, but also a robust, logically sound model for proactive, multi-echelon inventory distribution.

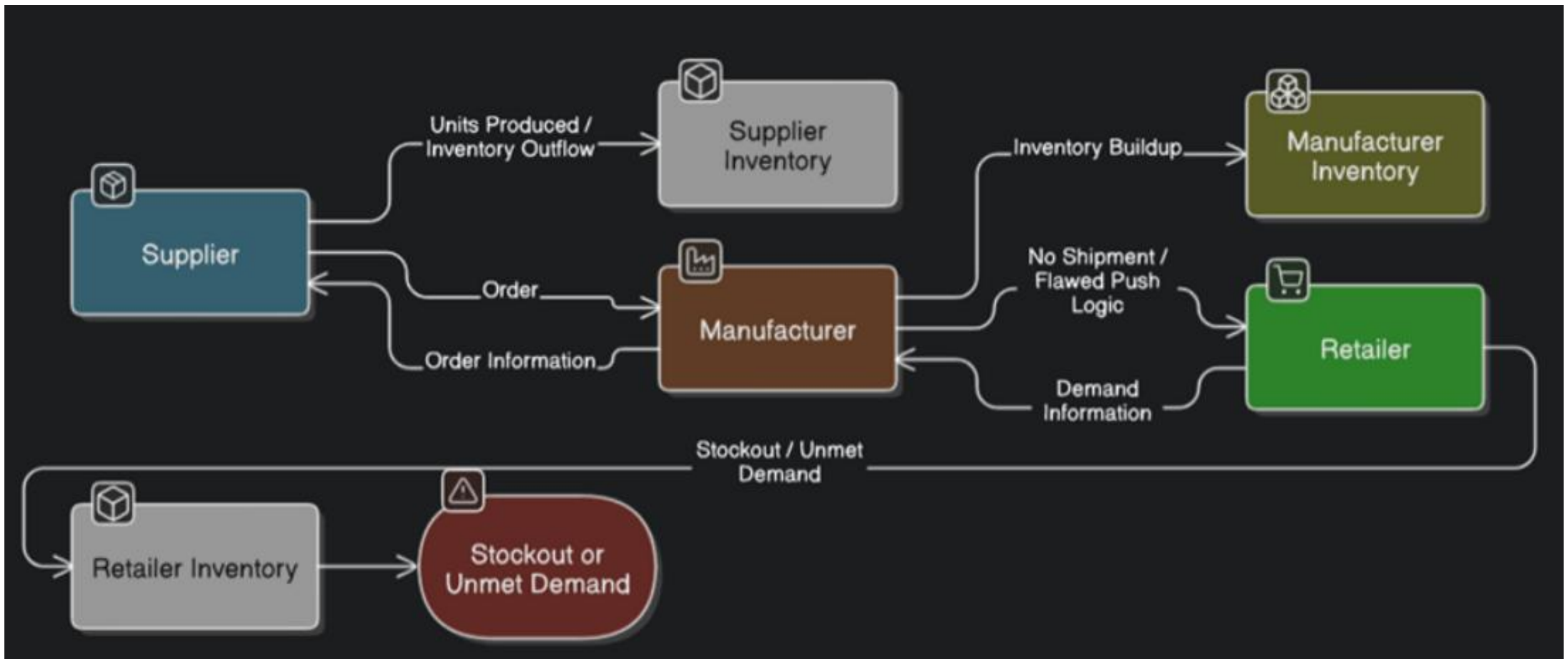

Figure 2: Conceptual Diagram of the "Hoarding Effect" Failure Mode. The Manufacturer (M) correctly places a large consolidated order with the Supplier (S) based on the Retailer's (R) needs, but a logical flaw prevents it from pushing inventory downstream, starving the Retailer.

## 4.3 The Strategic Choice Experiment: Validating the Final Framework

Leveraging the insights from our previous failures, we developed our final, successful collaborative model. This framework was then subjected to our main experiment, designed to test its ability to generate, evaluate, and quantify a portfolio of distinct strategic choices in response to a Transportation Disruption.

**4.3.1 Strategy Generation and Qualitative Evaluation:** Upon detecting the disruption, the Strategy Generation Agent (SGA) was tasked with formulating a response. The curated knowledge base for this experiment was designed specifically to test the SGA's ability to differentiate between distinct choices. It contained three specific, pre-defined tactical strategies modeled after common industry responses to logistics disruptions: a low-cost/high-risk option (Absorb Cost), a high-cost/low-risk option (Expedite Shipping), and a balanced alternative (Reroute Partial). The SGA successfully queried this knowledge base and generated these three distinct, viable strategies, as shown in Table 3. This result directly addresses RQ2, confirming the framework's ability to generate diverse mitigation plans from a curated set of options.

Table 3: Strategic Options Generated by the SGA for the Transportation Disruption

| **Strategy** | **Parameters** |
|---|---|
| Strategy 1 (Reroute Partial) | 'extra lead time': 2, 'transport cost premium': 75 |
| Strategy 2 (Expedite Shipping) | 'extra lead time': 0, 'transport cost premium': 200 |
| Strategy 3 (Absorb Cost) | 'extra lead time': 4, 'transport cost premium': 0 |

These three options were then passed to our 'VirtualExpertPanel'. The panel provided a qualitative rating for each strategy, presented in Table 4. The panel correctly identified the trade-offs, categorizing Strategy 3 as the low-cost/slow-speed option and Strategy 2 as the high-cost/high-speed option. This result confirms H1, demonstrating the feasibility of using an automated, AI-driven panel to qualitatively validate strategic choices.

Table 4: Qualitative Evaluation of Generated Strategies by the Virtual Expert Panel

| **Strategy** | **Cost Rating** | **Speed Rating** |
|---|---|---|
| Strategy 1 | Medium | Medium |
| Strategy 2 | Very High | Very Fast |
| Strategy 3 | Low | Very Slow |

**4.3.2 Quantitative Performance of Strategic Choices:** Finally, we performed a "what-if" analysis by running our now-stable simulation three times, once for each strategy. The quantitative performance of each choice is summarized in Table 5. All three strategies, executed within our robust collaborative framework, achieved a near-perfect service level, indicating a highly resilient system. The primary differentiator was the total cost.

Table 5: Quantitative Performance of Each AI-Generated Strategy

| **Strategy** | **Total Cost ($)** | **Service Level (%)** |
|---|---|---|
| Strategy 1 (Reroute Partial) | 16,896.00 | 100.0 |
| Strategy 2 (Expedite Shipping) | 19,767.50 | 100.0 |
| Strategy 3 (Absorb Cost) | 15,363.00 | 100.0 |

The inventory dynamics for each strategy are plotted in Figure 3, showing how each choice effectively mitigated the disruption period from day 60 to 80. Figure 4 visualizes the final trade-off, plotting the cost versus service level for each option. This plot represents the ultimate output of our framework: a clear, data-driven "efficient frontier" of strategic choices that can be presented to a human decision-maker. This successfully addresses RQ3 and provides a nuanced confirmation of H2, showing that our final framework is not just superior, but can generate a portfolio of superior choices.

## 5. Discussion

The results of our comprehensive computational experiments tell a story that is both cautionary and prescriptive. Our journey from a series of catastrophic system failures to a final, stable, and high performing model provides critical insights into the practical realities of deploying Generative AI within complex, dynamic business systems like supply chains. This section analyzes and interprets these findings, highlighting our key contributions to the theory and practice of AI in business analytics.

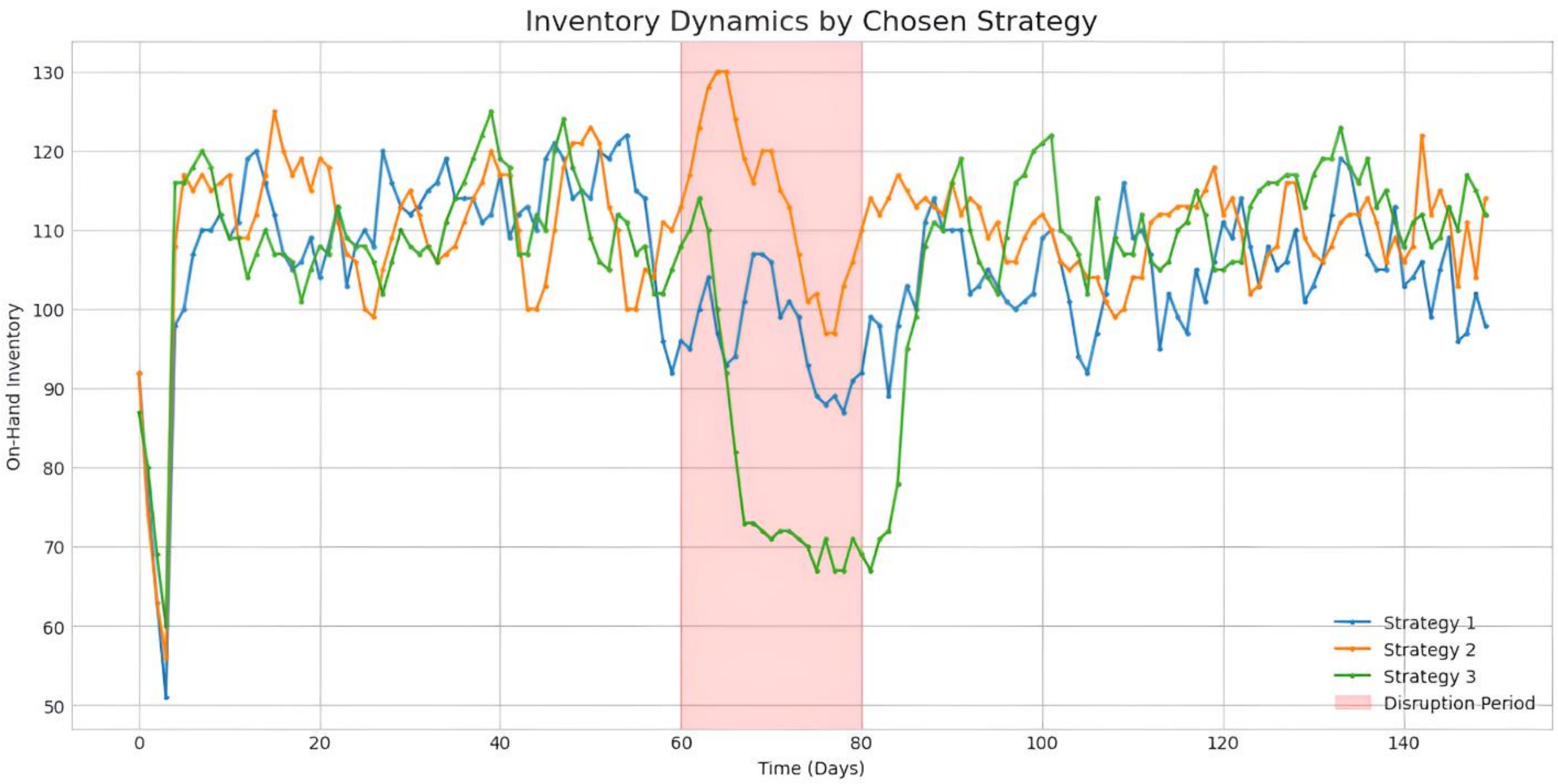

Figure 3: Inventory Dynamics of the Final Collaborative Model for each of the three SGA-generated strategies, showing effective mitigation of the disruption period (days 60-80)

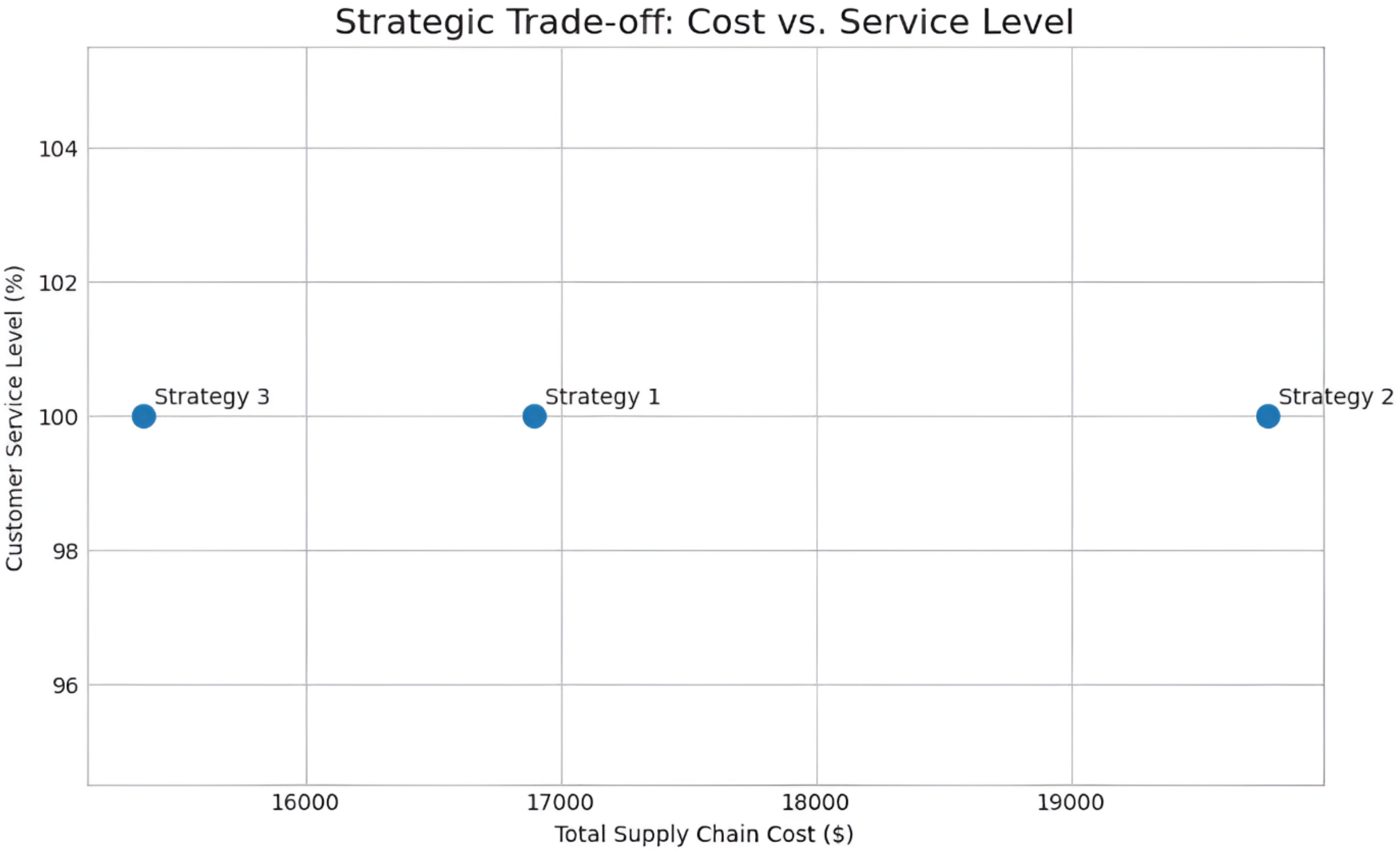

Figure 4: The Strategic Trade-off Frontier. This plot visualizes the final quantitative output, showing the cost and service level associated with each of the three viable strategies generated by the SGA.

## 5.1 The Collaboration Paradox: Why Naive AI Can Be Worse Than No AI

Our most striking initial finding was the repeated, spectacular failure of our non-collaborative and improperly implemented collaborative models. The Static Baseline, as expected, performed poorly, confirming that simple, fixed policies are inadequate in volatile environments. More surprisingly, our Model 2, the "Selfish RAG Agent," consistently performed no better, and in some cases worse, than this naive baseline. This finding presents a compelling paradox: endowing a single agent with advanced AI capabilities for local optimization did not translate to systemic improvement. Instead, its reactive, large-scale decisions, made without regard for system-wide inventory levels, often amplified the bullwhip effect, creating greater instability than the simple, predictable baseline.

Our findings here resonate with Fish et al., (2025), who observed LLMs autonomously develop supracompetitive strategies in pricing games. Our "selfish" AI agents similarly optimize locally, but in a cooperative setting, amplifying system-wide instability. Both studies reveal how unconstrained AI agents learn destabilizing behaviors directly from environmental incentives. This result serves as a powerful empirical demonstration of the dangers of deploying AI in a piecemeal fashion. It suggests that the behavior of an AI agent is not just a function of its own intelligence, but of the structure of the system in which it operates (Simon, 1962). Our early collaborative models (Model 3) further deepened this insight. The "hoarding effect" we observed was not a failure of the AI's strategic intent but a failure of the implemented operational protocol. These failures underscore a critical lesson: in tightly-coupled systems, the connections and protocols between agents are at least as important as the intelligence of the agents themselves.

## 5.2 The Synthesis of Strategy and Stability

The success of our final model (Model 4) was not achieved by simply adding more intelligence, but by creating a hierarchical system that synthesizes two distinct layers of AI-driven decision-making. This synthesis represents the core of our proposed framework.

The first layer is Proactive Strategic Policy-Setting. By tasking the SGA with setting systemwide inventory targets at the outset, we moved the AI's role from a reactive "firefighter" to a proactive "architect." The AI used a knowledge base of operations management principles to establish the global conditions for stability. This step was crucial, as it ensured the system was well-buffered from the start.

The second layer is Collaborative Operational Execution. The VMI-style information sharing and, critically, the proactive downstream replenishment policy, provided the operational stability needed to execute the AI's strategy. This layer tamed the bullwhip effect by replacing distorted, local order signals with a single, centralized view of system-wide needs. Our results prove that neither of these layers is sufficient on its own. A brilliant strategy is useless in an unstable system (as shown by our early failures), and a stable system with a poor strategy will still underperform. Resilience is an emergent property of their effective synthesis.

## 5.3 Implications for Business Analytics and the IEOM Community

Our findings have significant implications for the application of Generative AI in business analytics. First and foremost, they challenge the notion of "plug-and-play" AI solutions. Our research demonstrates that integrating an LLM-based agent into a business process is not merely a technical task but a deep exercise in systems design. To unlock the value of GenAI, firms may need to fundamentally re-architect their underlying operational processes to support the collaborative, data-sharing protocols that AI agents require to function effectively. While this represents a significant investment, the quantitative results from our final model demonstrate a compelling return; the computational overhead of the RAG agents is minimal compared to the immense financial losses from a single catastrophic stockout event, as evidenced by the performance of our baseline models (Table 2).

Second, our work highlights the continued, and perhaps even heightened, importance of classic Operations Research (OR) and management science principles in the age of AI. Our AI's success was contingent on its ability to access and apply foundational knowledge from inventory theory. This suggests that the most powerful applications of GenAI in business will not come from the models alone, but from a "human-in-the-loop" or "domain-knowledge-in-the-loop" approach where LLMs are grounded in the rigorous, time-tested knowledge of fields like OR (Lami, 2022).

Finally, our "Strategic Choice" experiment showcases a new potential role for AI in decision support. The framework's ability to generate, qualitatively evaluate, and quantitatively simulate a portfolio of viable strategies provides a powerful tool for human managers. It elevates the AI from a mere executor to a true strategic partner, one that can illuminate the trade-offs between different courses of action (e.g., the cost vs. service level frontier in Figure 4) and enhance the quality of human decision-making.

### 5.4 Societal Impact and Policy Implications

While our framework offers a pathway to more resilient supply chains, its deployment could have far-reaching societal consequences. Centralized, AI-driven collaboration, while improving efficiency, may disadvantage smaller suppliers who lack the technological capacity to integrate, potentially accelerating market consolidation and reducing supplier diversity. The role of human supply chain planners could shift from direct operational control to oversight and governance of AI systems (Acemoglu and Restrepo, 2018), creating both opportunities and displacement risks. Policymakers may need to address equitable supplier access, workforce retraining programs, and standards for safe AI–human collaboration to ensure that technological gains do not come at the expense of inclusivity and economic resilience.

### 5.5 Limitations and Future Research

While our study provides valuable insights, we acknowledge its limitations, which in turn motivate several exciting avenues for future research. Our simulation, though iteratively refined to be logically robust, remains a simplified representation of a real-world supply chain. Though minimalist, our environment shares key dynamics with real multi-echelon systems (e.g., decentralized information, lead times). Like the Beer Game (Oroojlooyjadid et al., 2020), it serves to isolate behavioral insights that are transferable to more complex networks, as evidenced by our discovery of the hoarding effect, which mirrors failures observed in practical VMI implementations (Disney and Towill, 2003). A critical next step is to address scalability by testing the framework in more complex, multi-product, multi-echelon network topologies. The “sim-to-real gap” is a significant hurdle, and future work must focus on real-world validation. This could involve partnerships with logistics firms to test the framework in a “shadow mode,” where it processes real-time data and recommends actions in parallel with human planners, allowing for a direct comparison of decision quality.

Furthermore, our ‘VirtualExpertPanel’ was a simplified proxy for qualitative evaluation. Future research could explore fine-tuning LLMs on extensive case study libraries to create more sophisticated expert personas. A particularly promising direction would be an ablation study to isolate the LLM’s value by comparing the RAG-based SGA against a traditional, rule-based strategy generator. Finally, an exciting avenue for research would be to integrate our strategic SGA with a tactical agent based on Reinforcement Learning (RL). The SGA could set the high-level policy (e.g., “prioritize service level”), which would then define the reward function for an RL agent that learns the optimal, fine-grained ordering and replenishment decisions. This hierarchical approach, combining the reasoning power of LLMs with the optimization power of RL, could represent the next step in creating truly intelligent and autonomous business systems.

## 6. Conclusion

This paper chronicled the journey to develop a resilient, AI-driven supply chain management system. Our research began with the hypothesis that a collaborative, intelligent agent framework would outperform traditional policies. While ultimately proven correct, our most significant findings emerged not from our final success, but from the insightful failures that preceded it.

Our central contribution is the empirical demonstration of the “collaboration paradox”: that intelligent, AI-driven systems can amplify instability if their operational protocols are not robust. We showed that even theoretically sound frameworks like VMI can fail if they do not account for critical execution logic, such as proactive downstream replenishment. The successful framework we present, which combines a proactive AI policy-setter with a stable, information-sharing execution model, serves as a blueprint for the necessary synthesis of strategic intelligence and operational stability.

Ultimately, this work provides both a novel “Strategic Choice” architecture and a cautionary tale. Our findings argue that the future of AI in business analytics lies not in creating monolithic AI oracles, but in architecting systems where AI agents are deeply integrated with the time-tested principles of operations research. By embracing and learning from the complexities of agent interaction, we can begin to unlock the true transformative potential of Generative AI.

## References

Acemoglu, D., & Restrepo, P. (2018). The race between man and machine: Implications of technology for growth, factor shares, and employment. *American Economic Review*, 108(6), 1488–1542. https://doi.org/10.1257/aer.20160696

Adadi, A., & Berrada, M. (2018). Peeking inside the black-box: A survey on explainable artificial intelligence (XAI). *IEEE Access*, 6, 52138–52160.

Agrawal, A., Gans, J., & Goldfarb, A. (2023). How large language models reflect human judgment. *Harvard Business Review*. https://ramaonhealthcare.com/how-large-language-models-reflect-human-judgment/

Agrawal, K. (2023). Towards adoption of generative AI in organizational settings. *Journal of Computer Information Systems*, 64, 1–16. https://doi.org/10.1080/08874417.2023.2240744

Calvano, E., Calzolari, G., Denicolò, V., & Pastorello, S. (2020). Artificial intelligence, algorithmic pricing, and collusion. *American Economic Review*, 110(10), 3267–3297. https://www.jstor.org/stable/26966472

Carbonneau, R., Laframboise, K., & Vahidov, R. (2008). Application of machine learning techniques for supply chain demand forecasting. *European Journal of Operational Research*, 184(3), 1140–1154.

Chopra, S., & Sodhi, M. S. (2004). Managing risk to avoid supply-chain breakdown. *MIT Sloan Management Review*, 46(1), 53–61.

Christopher, M., & Peck, H. (2004). Building the resilient supply chain. *International Journal of Logistics Management*, 15(2), 1–14.

Davenport, T. H., & Mittal, N. (2023). *All-in on AI: How smart companies win big with artificial intelligence*. Harvard Business Review Press.

Dhar, S., & Goh, M. (2025). Occam's razor in practice: Why simple, interpretable models win in business forecasting. Available at SSRN. https://doi.org/10.2139/ssrn.5404515

Disney, S. M., & Towill, D. R. (2003). The effect of vendor-managed inventory (VMI) dynamics on the bullwhip effect in supply chains. *International Journal of Production Economics*, 85(2), 199–215.

Fish, S., Gonczarowski, Y. A., & Shorrer, R. I. (2026). Algorithmic collusion by large language models. *arXiv preprint*, arXiv:2404.00806. https://arxiv.org/abs/2404.00806

Holland, J. H. (1992). *Adaptation in natural and artificial systems: An introductory analysis with applications to biology, control, and artificial intelligence*. MIT Press.

Hosseini, S., Ivanov, D., & Dolgui, A. (2019). Review of quantitative methods for supply chain resilience analysis. *Transportation Research Part E: Logistics and Transportation Review*, 125, 285–307. https://aquila.usm.edu/fac_pubs/16043

Klibi, W., Martel, A., & Guitouni, A. (2010). The design of robust value-creating supply chain networks: A critical review. *European Journal of Operational Research*, 203(2), 283–293.

Lami, O. (2022). Predictive and prescriptive analytics in operations management.

Lee, H. L., Padmanabhan, V., & Whang, S. (1997). The bullwhip effect in supply chains. *Sloan Management Review*, 38(3), 93–102.

Lewis, P., Perez, E., Piktus, A., Petroni, F., Karpukhin, V., Goyal, N., Küttler, H., Lewis, M., Yih, W. T., Rocktäschel, T., Riedel, S., & Kiela, D. (2020). Retrieval-augmented generation for knowledge-intensive NLP tasks. In *Advances in Neural Information Processing Systems*, 33.

Li, B., Mellou, K., Zhang, B., Pathuri, J., & Menache, I. (2023). Large language models for supply chain optimization. *arXiv preprint*, arXiv:2307.03875.

Makridakis, S., Spiliotis, E., & Assimakopoulos, V. (2018). Statistical and machine learning forecasting methods: Concerns and ways forward. *PLoS ONE*, 13(3), e0194889.

Oroojlooyjadid, A., Nazari, M. R., Snyder, L. V., & Takáč, M. (2020). A deep Q-network for the beer game: Deep reinforcement learning for inventory optimization. *Manufacturing & Service Operations Management*, 22(2), 294–312.

Ponomarov, S. Y., & Holcomb, M. C. (2009). Understanding the concept of supply chain resilience. *International Journal of Logistics Management*, 20(1), 124–143.

Raji, D., Denton, E., Bender, E. M., Hanna, A., & Paullada, A. (2021). AI and the everything in the whole wide world benchmark. In *Proceedings of the Neural Information Processing Systems Track on Datasets and Benchmarks*, 1.

https://datasets-benchmarks proceedings.neurips.cc/paper_files/paper/2021/file/084b6fbb10729ed4da8c3d3f5a3ae7c9-Paper-round2.pdf
Sheffi, Y. (2005). *The resilient enterprise: Overcoming vulnerability for competitive advantage*. MIT Press.
Silver, D., Schrittwieser, J., Simonyan, K., Antonoglou, I., Huang, A., Guez, A., Hubert, T., Baker, L., Lai, M., Bolton, A., Chen, Y., Lillicrap, T., Hui, F., Sifre, L., van den Driessche, G., Graepel, T., & Hassabis, D. (2017). Mastering the game of Go without human knowledge. *Nature*, 550(7676), 354–359.
Simon, H. A. (1962). The architecture of complexity. *Proceedings of the American Philosophical Society*, 106(6), 467–482.
Sterman, J. D. (1989). Modeling managerial behavior: Misperceptions of feedback in a dynamic decision-making experiment. *Management Science*, 35(3), 321–339. https://doi.org/10.1287/mnsc.35.3.321
Taleb, N. N. (2007). *The black swan: The impact of the highly improbable*. Random House.
Tang, C. S. (2006). Perspectives in supply chain risk management. *International Journal of Production Economics*, 103(2), 451–488.
Touvron, H., Martin, L., Stone, K., Albert, P., Almahairi, A., Babaei, Y., Bashlykov, N., Batra, S., Bhargava, P., Bhosale, S., et al. (2023). LLaMA 2: Open foundation and fine-tuned chat models. *arXiv preprint*, arXiv:2307.09288. https://arxiv.org/abs/2307.09288
Wang, S., Zhao, Y., Hou, X., & Wang, H. (2024). Large language model supply chain: A research agenda. *arXiv preprint*, arXiv:2404.12736. https://arxiv.org/abs/2404.12736
Wooldridge, M. (2009). *An introduction to multiagent systems* (2nd ed.). John Wiley & Sons.
Wu, J., & Sekiguchi, T. (2019). Applications of agent-based modeling and simulation in organizational behavior and human resource management: A bibliometric mapping study (2001–2019). Kyoto University Graduate School of Economics Discussion Paper No. E-19-008. https://www.econ.kyoto-u.ac.jp/dp/papers/e-19-008.pdf

## Biography

**Soumyadeep Dhar** is a final-year student in the Department of Industrial and Systems Engineering at the Indian Institute of Technology Kharagpur, where he is pursuing an integrated Bachelor's and Master's degree. His research interests lie at the intersection of operations management, artificial intelligence, and optimization. His work has been accepted at international conferences such as the Genetic and Evolutionary Computation Conference (GECCO), and his manuscripts are currently under review at leading journals including *Applied Energy*. He will join the Stanford Graduate School of Business as a Predoctoral Research Fellow in the Operations, Information, and Technology (OIT) area.

## Acknowledgements

We would like to express our sincere gratitude to all the research scholars associated with the Operations Research and Data Science (ORDS) laboratory at IIT Kharagpur for their timely feedback and support.